\documentclass[letterpaper]{article}
\usepackage{aaai}
\usepackage{times}
\usepackage{helvet}
\usepackage{courier}
\usepackage{url}
\usepackage{booktabs}
\usepackage{multirow}
\usepackage{graphicx}
\usepackage{url}

\nocopyright

\newcommand{\citet}[1]{\citeauthor{#1} (\citeyear{#1})}

\begin{document}
%
\title{Modeling Human Temporal Uncertainty in Human-Agent Teams}
\author{Maya Abo Dominguez\thanks{Authors listed alphabetically but contributed equally.},
William La\footnotemark[1],
James C. Boerkoel Jr.\\
Human Experience and Agent Teamwork Lab (heatlab.org)\\
Harvey Mudd College\\
Claremont, California 91711\\
\{mabodominguez, wla, boerkoel\}@hmc.edu
}
\maketitle
\begin{abstract}
\begin{quote}
Automated scheduling is potentially a very useful tool for facilitating efficient, intuitive interactions between a robot and a human teammate.
However, a current gap in automated scheduling is that it is not well understood how to best represent the timing uncertainty that human teammates introduce.
This paper attempts to address this gap by designing an online human-robot collaborative packaging game that we use to build a model of human timing uncertainty from a population of crowdworkers.
We conclude that heavy-tailed distributions are the best models of human temporal uncertainty, with a Log-Normal distribution achieving the best fit to our experimental data.
We discuss how these results along with our collaborative online game will inform and facilitate future explorations into scheduling for improved human-robot fluency.

\end{quote}
\end{abstract}

\section{Introduction} 

As use of robots to assist humans gains popularity in a variety of settings, the fluency and intuitiveness of human-robot interactions becomes more important. A key challenge in any teamwork is to ensure that all team members have an accurate model of how/when other agents might complete their action.
This is particularly important in close-collaborative settings such as manufacturing, where the fluidity and effectiveness of interactions depends on a human's ability to anticipate the actions of their robot counterpart, and \textit{vice versa}. 

An example of this is a human working in a fulfillment center packing boxes while robots deliver pallets of items that need to be packaged and shipped.
These robots need to be able to anticipate when to bring the items to the human packers so that they can achieve a smooth workflow packing the items: if the robots are not timed correctly, there is a risk that the human coworker will become either over- or under-whelmed by the pace.
A primary challenge is that humans' volatility introduces scheduling uncertainty from the perspective of the robot.

\begin{figure}
    \centering
    \includegraphics[width = 200pt]{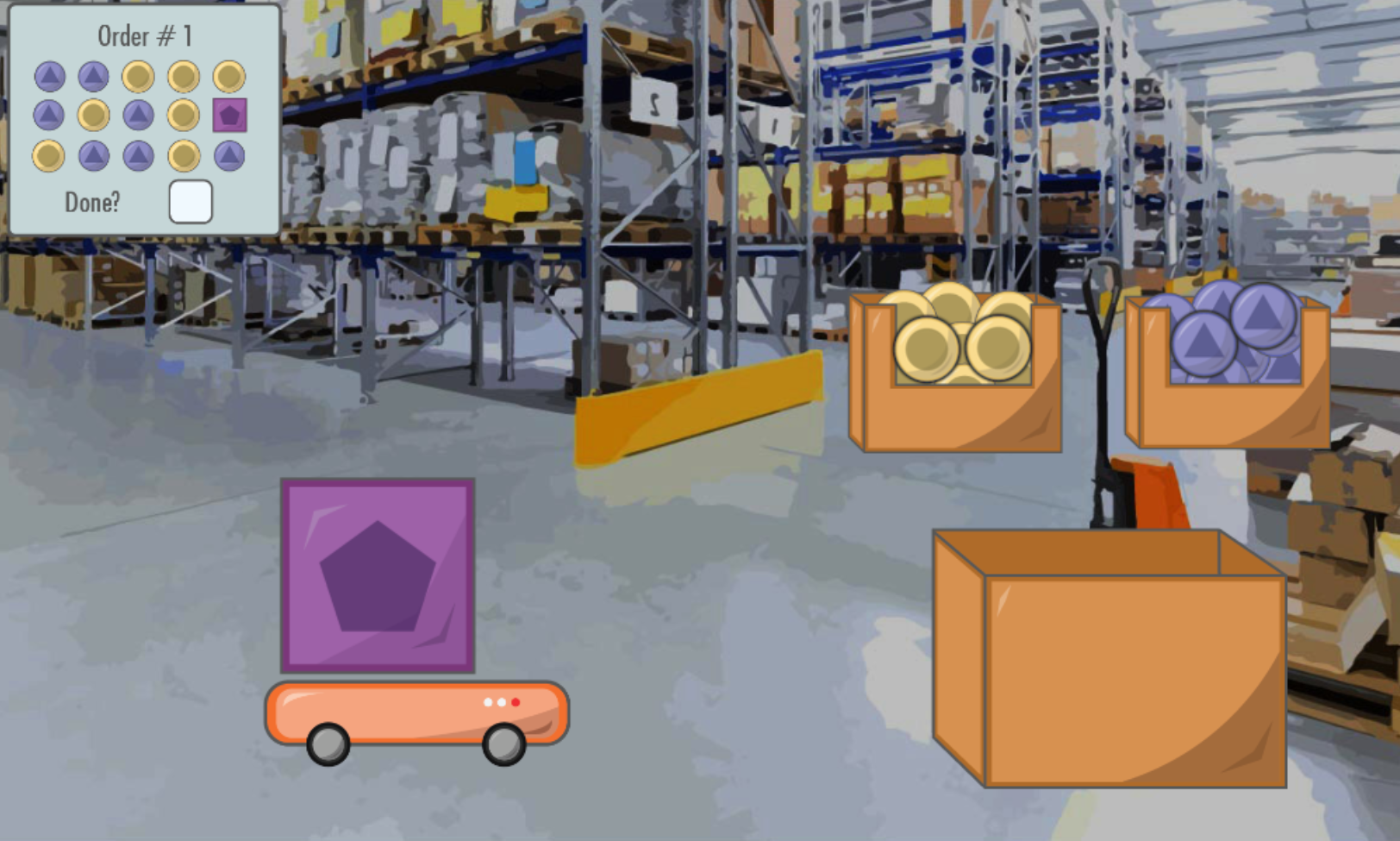}
    \caption{A screenshot of the game used as the interactive task. The robot is shown in the lower left of the screen carrying a purple box. Players must put items into the box in the sequence specified by the order card in the top left corner. A video of the gameplay can be found at:\\ \url{https://youtu.be/XOOkQFKq9NE}}
    \label{fig:gameplay}
\end{figure}

This motivates the primary investigation of this paper: how to best model the temporal uncertainty that humans introduce to human-robot interactions in a way that can be leveraged in the automated scheduling of robots? 
This paper starts by explaining recent work on using automated scheduling for human-robot teamwork.
Next, the primary contribution of this paper is a new online human-robot collaborative packaging task shown in Figure \ref{fig:gameplay} that can be used to both gather information about human-sources of timing in collaborative tasks and also facilitate evaluating further hypotheses related to human-robot fluency.
We then discuss an exploratory experiment that we ran on Amazon Mechanical Turk to gather timing data across a population of crowdworkers.
We use the results to build empirical distributions of the timing of human actions and evaluate how well various candidate probability density functions fit the data.
We conclude by discussing how these models of human temporal uncertainty inform a variety of next directions and hypotheses in our exploration of using automated scheduling to improve human-robot fluency.


\section{Background} 
Scheduling when robots execute their actions is important for the efficacy, efficiency, and fluency of human-robot teams \cite{hoffman2014timing}.
Recent work characterizes how temporal networks are an efficient way to represent the scheduling problems of human-robot teams that supports both reasoning over the constraints and also monitoring execution so that scheduling advice can be dispatched in real-time \cite{castro2017takes,maniadakis2016collaboration,maniadakis2017time}.
As explained in more detail by \citet{castro2017takes}, a challenge, of course, is that humans pose an added source of variability and unreliability when it comes to scheduling their actions.
\citeauthor{castro2017takes} propose two approaches to representing the scheduling uncertainty that human's introduce---(1) bounding human sources of timing uncertainties to fixed intervals and (2) modeling human sources of timing uncertainty as a continuous probability (and/or cumulative) density functions (pdf/cdf).
This work addresses the second of the proposals by designing an experiment to determine the best model of human sources of timing uncertainty. 



\section{Experimental Design}


The goal of the experiment was to create a task that virtually simulates a realistic human-robot interaction and allows us to capture human sources of timing uncertainty.
To that end, we created a web-game based on a human-robot team that must work together to package and fulfill orders. 




\begin{figure*}[ht]
    \centering
    \includegraphics[width = 6.5in, trim=0 7 0 7, clip]{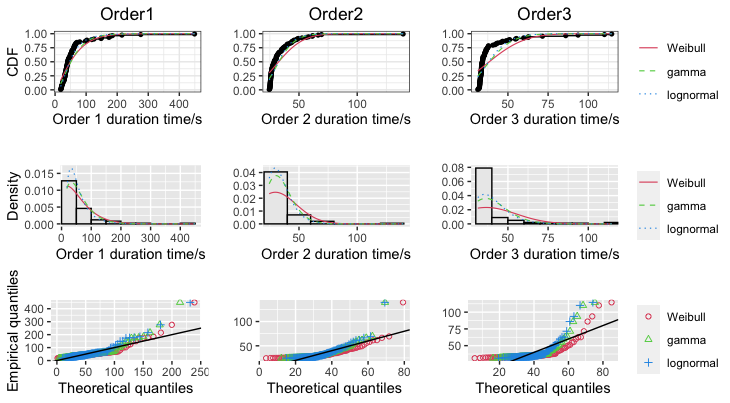}
    \caption{Graphical model comparisons for Weibull, gamma and Log-Normal distributions. The first row consists of CDF plots, second row are model density plots against the data histograms and the third row are Q-Q plots.} 
    \label{fig:plots}
\end{figure*}


\subsection{Collaborative Packaging Task}
Players must cooperate with the robot to complete a total of three orders to complete the game. 
As shown in Figure \ref{fig:gameplay}, each order presented to the player contains a specific sequence of items in which the package must be packed. 
The player ``packs" each box by dragging and dropping items into the box.
Some items can be fetched directly by the player from nearby stationary bins while other items must be brought into the playing area by the robot. 
If players attempt to place an item in the box in an incorrect order, the item simply fails to pack and returns to the staging area. 
Each order is completed when the box contains the correct sequence of items and the player clicks ``send". 

To simplify the flow of the game and reduce confusion for the player, the robot's schedule is set beforehand.
However, the robot waits for the player to pick up the object it is carrying before leaving the meeting point. 
Timing data was collected through internal timers within the game. 
To capture the duration of each order and the interaction as a whole, a timer would start once the order had begun and run until the order had been completed.
Once all three orders are completed, participants were thanked and asked to complete a post-game survey drawn from previous work on evaluating fluency \cite{hoffman2019evaluating}, which will inform our future research directions.
For a more detailed explanation of game play and footage of the game, see our instructional video (\url{https://youtu.be/XOOkQFKq9NE}). 

\subsection{Amazon Mechanical Turk}
We recruited participants for our study through Amazon Mechanical Turk (MTurk), a crowdsourcing website for virtual tasks. 
Past work has shown that data collected from MTurk workers compare well to data collected from traditional, in-person human experiments \cite{mason2012mturk}.
100 participants were recruited from MTurk and were each compensated \$1.25 for each task, which took approximately 5 minutes to complete. 

We had each MTurk worker play through all three orders of our packing game and then asked them to fill out the post-game survey. 
The within-subjects design allows us to directly assess how, if at all, the nature of human timing uncertainty (i.e., shape of the distribution) evolves with learning effects over time.
It also allows us to examine whether different patterns emerge when we consider micro (i.e., individual orders) vs. macro (the entire interaction) human-robot interactions.

All participants were verified MTurk workers over the age of 18 and residing in the United States, and had an approval rating of 99\% or above for previous tasks. 
We excluded results from MTurk workers that did not fully complete our game, completed our game more than once, or did not complete the followup survey.



\section{Results} 
Once the timing data was collected from the experiment, we created empirical distributions for how long it took players to complete each of the 3 orders as well as the to complete the overall game.
Indeed, as can be seen in the histograms in the middle row of Figure \ref{fig:plots}, the timing data generally points to a heavy-tailed distribution.
A possible reason for the heavy-tailed nature of our data is that for each order there was a lower bound on how quickly the order could be completed due to the pre-scheduled actions of the robot, and thus we would expect to see a congregation of data points near the minimum time, but none below it. 
Further, this lower bound was determined by how long the worker had to wait for the robot to deliver its item to be packaged rather than the complexity of the task itself, with order 3 having the longest wait time.
Conversely, there were no time constraints dictating how long participants could take, so the longer tail was likely composed of a combination of slower, more careful workers and workers who needed extra time to recover from a misunderstanding or mistake.

In addition, the variance of the first order is much greater than that of orders 2 and 3, even though Order 1 was the ``simplest" order.
This is likely due to learning effects---workers are unfamiliar with the game controls and objective during the first order.
The decrease in variance in orders 2 and 3 reflects that users quickly learned the game. 
These trends can all be verified in Table \ref{fig:table2}, where we report the mean and standard deviation of our data in line 1 for the Normal and Log-Normal distributions that best fit our timing data.

\begin{table*}
    \centering
    \begin{tabular}{ccccccccc}
        \toprule
        Distributions & \multicolumn{2}{c}{Order 1} & \multicolumn{2}{c}{Order 2} & \multicolumn{2}{c}{Order 3} & \multicolumn{2}{c}{Overall} \\
        \cmidrule(lr){1-1} \cmidrule(lr){2-3} \cmidrule(lr){4-5} \cmidrule(lr){6-7} \cmidrule(lr){8-9}
         Normal & $\overline{x}$ = 60.04 & $\sigma$ = 59.57 & $\overline{x}$ = 34.04 & $\sigma$ = 14.56 & $\overline{x}$ = 39.58 & $\sigma$ = 14.77 & $\overline{x}$ = 133.66 & $\sigma$ = 74.01 \\
         Weibull & $\gamma$ = 1.30 & $\alpha$ = 65.97 & $\gamma$ = 2.28 & $\alpha$ = 38.21 & $\gamma$ = 2.56 & $\alpha$ = 44.3 & $\gamma$ = 1.95 & $\alpha$ = 151.59 \\
         Gamma & $\gamma$ = 2.18 & $\beta$ = 0.04 & $\gamma$ = 9.39 & $\beta$ = 0.28 & $\gamma$ = 11.88 & $\beta$ = 0.30 & $\gamma$ = 5.45 & $\beta$ = 0.95 \\
         Log-Normal & $\overline{x}$ = 3.85 & $\sigma$ = 0.62 & $\overline{x}$ = 3.47 & $\sigma$ = 0.30 & $\overline{x}$ = 3.64 & $\sigma$ = 0.26 & $\overline{x}$ = 4.80 & $\sigma$ = 0.39 \\
        \bottomrule
    \end{tabular}
    \caption{Parameters for the fitting distributions. In the table, $\overline{x}$ symbolizes the mean, $\sigma$ is the standard deviation, $\gamma$ is the shape, $\alpha$ is the scale, and $\beta$ is the rate of their respective distributions.} 
    \label{fig:table2}
\end{table*}
\begin{table*}
    \centering
    \begin{tabular}{ccccccccccccc}
        \toprule
        \multirow{2}{*}{Distributions} & \multicolumn{3}{c}{Order 1} & \multicolumn{3}{c}{Order 2} & \multicolumn{3}{c}{Order 3} & \multicolumn{3}{c}{Overall}\\
        \cmidrule(lr){2-4} \cmidrule(lr){5-7} \cmidrule(lr){8-10} \cmidrule(lr){11-13} 
        {} & AD & AIC & BIC & AD & AIC & BIC & AD & AIC & BIC & AD & AIC & BIC \\
        \midrule
         Normal & 12.9 & 1105.2 & 1110.4  & 10.1 & 823.5 & 828.7 & 17.4 & 826.3 & 831.5 & 10.2 & 1148.6 & 1153.8 \\
         Weibull & 6.7 & 1010.2 & 1015.5 & 10.3 & 811.0 & 816.2 & 17.1 & 822.8 & 828.0 & 8.33 & 1116.7 & 1121.9 \\
         Gamma & 5.6 & 994.6 & 999.8 & 6.9 & 762.0 & 767.2 & 14.9 & 770.2 & 775.4 & 5.9 & 1084.5 & 1089.7 \\
         Log-Normal & \textbf{2.6} & \textbf{961.9} & \textbf{967.1} & \textbf{5.8} & \textbf{739.5} & \textbf{744.7} & \textbf{13.6} & \textbf{745.7} & \textbf{751.0} & \textbf{4.0} & \textbf{1061.0} & \textbf{1066.3} \\
        \bottomrule
    \end{tabular}
    \caption{Statistical test results for our data fitted to the chosen density distributions.}
    \label{fig:table}
\end{table*}

Next, we examine which theoretical probability distribution best modeled our data.
A model was obtained by fitting our data to several different density distributions using the maximum likelihood estimation (MLE) functionality in the R package  \verb|fitdistrplus| \cite{delignette-muller2015fitdistrplus}.
After exploring a variety of distribution options, we converged on the heavy-tailed distributions of Weibull, Gamma and Log-Normal as providing the most likely explanation of the timing data.
We also include the Normal distribution for reference, despite its relatively poor fit.
Table \ref{fig:table2} reports the parameters that achieved each of the respective MLEs.

We demonstrate the quality of the fit of these three MLE distributions to our empirical data in two ways.
First, Figure \ref{fig:plots} shows these fits visually.
The top row contains the empirical vs. predicted cumulative density function, the second row contains the empirical histogram vs. theoretically predicted probability density function, and the third row contains a Q-Q plot which show the predicted vs. empirical quantiles of the data.

The second way we assess fit is through statistical measures provided by the \verb|fitdistrplus| package , which includes the Anderson-Darling (AD) statistic, the Akaike Information Criterion (AIC), and Bayesian Information Criterion (BIC). 
Table \ref{fig:table} reports these fit metrics for each order across the four distributions we evaluated. 
The AD statistic is related to the probability that a given data set is \textit{not} drawn from the same distribution as a model, so the lower the value, the more likely the data set \textit{is} drawn from the same distribution as the model \cite{Anderson2011}. 
The AIC and BIC are closely related metrics of quality of fit.
Loosely, the AIC is a measure from information theory that measures the amount of information is lost by adopting a statistical model of a process rather then empirical data itself, and minimizing the BIC corresponds to maximizing the posterior model probability \cite{wit2012AICBIC}.
For all three criteria, a lower value indicates a better model fit. 
From the statistics observed in Table \ref{fig:table}, the Log-Normal density distribution provides the best fit to our our empirical data across all orders. 
Interestingly, while the parameters of the Log-Normal density shift across orders due to variance in timings and learning effects, the fact Log-Normal provides the best fit is robust across interactions.
The Log-Normal distribution is one where the \emph{logarithm} of the random variable involved is normally distributed.
So compared to a Normal distribution, the variance on the left side of the mode gets exponentially compressed, whereas on the other side it exponentially expands.

Our results corroborate those of \citeauthor{blenn2016human}, who found that Log-Normal distribution accurately captures the timing of human-human interactions in online social networks. 
Interestingly, many other well known models regarding human timing in HCI also yield logarithmic relationships such as target acquisition time is logarithmically related to distance (Fitt's Law), the time to make decision is logarithmitically related to the number of options (Hick's law) and the time it takes to execute a task decreases logarithmically with practices (Power Law of Practice).


\section{Related Work and Next Directions}


Temporal networks have have recently been augmented to better facilitate human-robot collaboration \cite{maniadakis2016collaboration,maniadakis2017time}.
This, in turn, has facilitated more detailed explorations into how the precise timing of robot actions influences human's behaviors.
For instance, \citet{isaacson2019mad} explore how temporal networks can be used to measure key aspects of human-robot fluency, which \citet{hoffman2019evaluating} defines as ``coordinated meshing  of  joint  activities  between  members  of  a  well-synchronized team".
Further, \citet{castro2017takes} show that temporal networks can be used to drive the scheduling of robot actions in a way that can improve overall human-team efficiency when the robot starts the interaction, which corroborates previous studies showing that scheduling anticipatory actions can improve team fluency \cite{hoffman2007effects}.

Similar to the goals of the previous fluency work, the game design proposed by this paper aims to provide a framework to allow precise measurement of fluency metrics, as well as provide an avenue to implement and test automated scheduling, within the context of a virtual human-robot interaction. 
More specifically, our framework is extensible in a way that allows the capturing of more nuanced timing. 
For instance, we built-in a series of timers triggered by contingent events to capture, for example, the delay between the robot bringing the item and the player ``picking it up". 
Further, due to the modular nature of the game other aspects of the interaction could also be timed.
In the next section, we outline how our online framework could used to run large-scale, online experiments that further explore human-robot team fluency.

\section{Conclusions and Future Research Directions}
In this paper we address an open question about how best to modeling  human  sources  of  timing  uncertainty as a continuous probability \citet{castro2017takes}.
We did this by creating a framework for collecting human temporal data through a web-based game which simulates the collaborative human-robot teamwork found in a fulfillment center. 
We used this framework to run an exploratory, online experiment that allowed us to capture key timing information regarding the durational uncertainty of human teammates using crowdworkers on Amazon Mechanical Turk.
We collected and fit data from our MTurk exploration to well-known density distributions, discovering that the heavy-tailed Log-Normal distribution provides the best fit---i.e., the most likely explanation.
Further, these findings were robust across learning effects, with Log-Normal providing the best fit across all orders, and at both the micro and macro levels, as exhibited by Log-Normal providing the best fit both for individual orders and across the duration of the entire interaction.

In the future, we hope to employ our experimental framework to facilitate two major research directions.
First, we can exploit the fact that our framework was designed to capture precising timing data in order to capture and evaluate previously proposed fluency metrics \cite{hoffman2007effects,hoffman2019evaluating,isaacson2019mad}.
The fact that our experimental framework can be scaled across large populations of crowdworkers while providing precise timing and fluency measures can provide an important tool for exploring fluency and potentially providing important insights in creating more fluent human-robot interaction in the real world.

Second, we can apply our insights regarding the nature of human-timing uncertainty to improve the dynamic scheduling of robotic tasks within human-robot teamwork.
For instance, we can use our Log-Normal insights to schedule robot sub-tasks using the latest approaches for scheduling under uncertainty that maximize  likelihood of success \cite{akmal2020quantifying}, which we in turn hypothesize will also maximize  participants' qualitative impressions of fluency \cite{hoffman2019evaluating}.
If our model is valid, we would expect the human users to have a more positive impression of the interaction, specifically with their perception of the fluency, efficiency, and pacing of the game.
We are also interested in validating whether or not Log-Normal are accurate models of human timing uncertainty in physical human-robot interactions.
Finally, we are interested in investigating whether or not we can learn trends across learning effects that allow us to quickly adapt robot scheduling according to the current pace of a particular user interaction.

\section{Acknowledgements}
Funding for this work was graciously provided by the National Science Foundation under grant IIS-1651822 and the Rose Hills Foundation. Thanks to the anonymous reviewers, HMC faculty, staff, and fellow HEATlab members for their support and constructive feedback, with a special thanks to Hannah Davalos, Ryan Martinez, and Vibha Rohilla for their extra efforts in supporting this work. 

\bibliographystyle{aaai}

\bibliography{references}

\end{document}